\documentclass[superscriptaddress, nofootinbib, prd, 12pt]{revtex4-1}
\usepackage[utf8]{inputenc} 
\usepackage[T1]{fontenc}    
\usepackage{hyperref}       
\usepackage{url}            
\usepackage{booktabs}       
\usepackage{amsfonts}       
\usepackage{nicefrac}       
\usepackage{microtype}      
\usepackage{xcolor}         
\usepackage{amsmath}
\usepackage{amssymb}
\usepackage{graphicx}
\usepackage{pgfplots}
\usepackage{subcaption}
\usepackage{wrapfig}
\usepackage{subcaption}

\newcommand{\ND}{N_\mathcal{D}}
\newcommand{\Loss}{\mathcal{L}}
\newcommand{\muL}{\mu_{\mathcal{L}}}
\newcommand{\sigmaL}{\sigma_{\mathcal{L}}}
\newcommand{\relvar}{\epsilon_{\mathcal{L}}}
\newcommand{\erf}{{\rm erf}}
\newcommand{\bK}{\mathbf{K}}
\newcommand{\bNTK}{\boldsymbol{\Theta}}
\newcommand{\zt}{\widetilde{z}^{(L)}}

\begin{document}

\title{Uncertainty Quantification From Scaling Laws in Deep Neural Networks}

\author{Ibrahim Elsharkawy}
\email{ie4@illinois.edu}
\affiliation{Department of Physics, University of Illinois Urbana-Champaign, Urbana, IL, USA}
\author{Benjamin Hooberman}
\email{benhoob@illinois.edu}
\affiliation{Department of Physics, University of Illinois Urbana-Champaign, Urbana, IL, USA}
\author{Yonatan Kahn}
\email{yf.kahn@utoronto.ca}
\affiliation{Department of Physics, University of Illinois Urbana-Champaign, Urbana, IL, USA}
\affiliation{Department of Physics, University of Toronto,
  Toronto, ON, Canada}
  \affiliation{Vector Institute,
  Toronto, ON, Canada}

\begin{abstract}
Quantifying the uncertainty from machine learning analyses is critical to their use in the physical sciences. In this work we focus on uncertainty inherited from the initialization distribution of neural networks. We compute the mean $\muL$ and variance $\sigmaL^2$ of the test loss $\Loss$ for an ensemble of multi-layer perceptrons (MLPs) with neural tangent kernel (NTK) initialization in the infinite-width limit, and compare empirically to the results from finite-width networks for three example tasks: MNIST classification, CIFAR classification and calorimeter energy regression. We observe scaling laws as a function of training set size $\ND$ for both $\muL$ and $\sigmaL$, but find that the coefficient of variation $\relvar \equiv \sigmaL/\muL$ becomes independent of $\ND$ at both infinite and finite width for sufficiently large $\ND$. 
This implies that the coefficient of variation of a finite-width network may be approximated by its infinite-width value, and may in principle be calculable using finite-width perturbation theory.
\end{abstract}

\maketitle

\section{Introduction}

Deep learning techniques have improved performance beyond conventional methods in a wide variety of tasks. However, for neural networks in particular, it is not straightforward to assign network-induced uncertainty on their output as a function of network architecture, training algorithm, and initialization~\cite{Chen:2022pzc}. One approach to uncertainty quantification (UQ) is to treat any individual network as a draw from an ensemble, and identify the systematic uncertainty with the variance in the neural network outputs over the ensemble~\cite{osband2016deep,lakshminarayanan2017simple}. This variance can certainly be measured empirically by training a large ensemble of networks, but it would be advantageous to be able to predict it from first principles. This is possible in the infinite-width limit of multi-layer perceptron (MLP) architectures, where the statistics of the network outputs after training are Gaussian with mean and variance determined by the neural tangent kernel (NTK)~\cite{jacot2018neural,arora2019exact,lee2019wide}. For realistic MLPs with large but finite width $n$, one can compute corrections to this Gaussian distribution that are perturbative in $1/n$~\cite{Roberts:2021fes}. However, the finite-width computation depends on tensors of size $\ND^4$, where $\ND$ is the training set size, which rapidly becomes unwieldy as $\ND$ grows large ~\cite{Chen:2022pzc,osband2016deep,lakshminarayanan2017simple}.

To sidestep the difficulties of large $\ND$, we propose to exploit scaling laws~\cite{kaplan2020scaling,bahri2021explaining,bordelon2024dynamical}, an apparently ubiquitous phenomenon in deep learning where test loss $\Loss$ follows a power law $\Loss \propto \ND^{-\alpha_D}$ with some task-dependent scaling exponent $\alpha_D$. Such scaling laws have been observed for both finite-width and infinite-width networks, and have been argued to be relatively insensitive to the neural network architecture for a given task, notably including whether or not a network is at finite width. Without attempting to explain the origin of these scaling laws, we can nonetheless use their existence and the simplicity of the infinite-width Gaussian statistics to attempt to extrapolate the expected mean test loss $\mu_L$ and its variance $\sigmaL^2$ to large $\ND$. The infinite-width prediction only depends on matrices of size $\ND \times \ND$, making a comparison to experiments more feasible. 

In this work, we compute $\muL$ and $\sigmaL^2$ for infinite-width MLPs for regression tasks on three example datasets: the benchmark MNIST classification problem formulated as a regression to 1-hot labels, the benchmark CIFAR classification problem also formulated as a regression to 1-hot labels, and an example from high-energy physics (HEP), energy calibration in a calorimeter detector. For all examples, we find nontrivial scaling laws for both $\muL$ and $\sigmaL^2$, but with related scaling exponents such that the coefficient of variation
\begin{equation}
    \label{eq:relvardef}
    \relvar \equiv \frac{\sigmaL}{\muL}
\end{equation}
asymptotes to a scaling exponent of $\alpha_D \approx 0$ at sufficiently large $\ND$. We give a plausibility argument for this somewhat surprising ``invariant'' of infinite-width architectures, and we demonstrate with numerical experiments that very similar scaling exponents persist at finite width.

There is a large body of work investigating scaling laws for wide networks and linear models, including Refs.~\cite{bordelon2020spectrum,canatar2021spectral,atanasov2022onset,bordelon2024dynamics,bahri2021explaining,Maloney:2022cvb,adlam2023kernel,bordelon2024dynamical,Zhang:2024mcu,bordelon2024featurelearningimproveneural}. To our knowledge, our work is the first to demonstrate the scaling exponent $\alpha_D \approx 0$ for the coefficient of variation $\relvar$, in the NTK parameterization where finite-width corrections may in principle be computed perturbatively. In future work we intend to study how our results interplay with related work in mean-field or $\mu$-parametrization~\cite{bordelon2024dynamics}, as well as to what extent feature learning is important for the effects of finite width on the coefficient of variation.

\section{Problem setup}
\label{sec:problem}

In order to maintain theoretical control and confine all stochasticity to the initialization distribution, we work with MLPs in critically initialized NTK parametrization, trained with full-batch gradient descent (GD). We have shown in \ref{app:adam} that our results generalize to other optimizing schemes, such as ADAM. The MLP forward pass is given by
%
\begin{equation}
z_i^{(1)} = b_i^{(1)} + \sum_{j=1}^{n_0} W_{ij}^{(1)} x_j, \quad z_i^{(\ell+1)} = b_i^{(\ell+1)} + \sum_{j=1}^{n_\ell} W_{ij}^{(\ell+1)} \sigma\left(z_j^{(\ell)}\right),
\end{equation}
where $x_j$ are the components of an input vector $\vec{x} \in \mathbb{R}^{n_0}$, $W_{ij}^{(\ell)}$ and $b_i^{(\ell)}$ are weights and biases at layer $\ell$, $\sigma(z)$ is the activation function, all $L-1$ hidden layers have the same width $n$, and the network output is a vector $\vec{z}^{(L)} \in \mathbb{R}^{n_L}$. To facilitate analytic computations of the NTK (see Sec.~\ref{sec:NTK} below), we choose $\sigma(z) = \erf(z)$. For this choice of activation, initializing biases to zero and drawing weights in layer $\ell$ from zero-mean Gaussian distributions with variance $1/n_{\ell-1}$
enforces criticality, namely that the typical magnitude of preactivations does not grow or shrink exponentially with depth~\cite{Roberts:2021fes}. Each network is then trained with full-batch gradient descent,
\begin{equation}
 \theta_\mu \leftarrow \theta_\mu - \eta \sum_{\nu} \lambda_{\mu \nu} \frac{d\mathcal{L}_\mathcal{A}}{d\theta_\nu}, \qquad \mathcal{L}_\mathcal{A} \equiv \frac{1}{n_L\ND} \sum_{\alpha \in \mathcal{A}}\frac{1}{2} || \vec{z}^{(L)}_\alpha - \vec{y}_\alpha||^2,
\label{eq:GDUpdate}
\end{equation}
where $\Loss_{\mathcal{A}}$ is the MSE loss on the training set $\mathcal{A} = \{\vec{x}_\alpha, \vec{y}_\alpha \}$, $\theta_\mu$ indexes the weights and biases, $\eta$ is the global learning rate, and (following~\cite{Roberts:2021fes}) the learning rate tensor $\lambda_{\mu \nu}$ is
\begin{equation}
\label{eq:lambdadef}
\lambda_{b^{(\ell)}_{i_1} b^{(\ell)}_{i_2}} = \delta_{i_1 i_2}\lambda_b/{\ell}, \qquad \lambda_{W^{(\ell)}_{i_1 j_1}W^{(\ell)}_{i_2 j_2}} = \delta_{i_1 i_2}\delta_{j_1 j_2} \lambda_W/n.
\end{equation}

All of the above scalings permit a sensible infinite-width limit, $n \to \infty$. Our free hyperparameters at infinite width are therefore $\eta$, $L$, and $\lambda_b/\lambda_W$. We specify particular choices for these hyperparameters in Sec.~\ref{sec:experiments}, but have checked that our results do not appear to depend on these choices; see Sec.~\ref{sec:LearningRateRatio} for a discussion of the dependence of $\relvar$ on the learning rate ratio $\lambda_b/\lambda_W$. We are interested in $\muL \equiv \mathbb{E}[\mathcal{L}_{\mathcal{B}}]$ and  $\sigmaL^2 \equiv \text{Var}[\mathcal{L}_B]$ as a function of $\ND$, where $\mathcal{B}$ is the test set and the expectation and variance are taken over the initialization distribution.


\section{Infinite-width predictions}
\label{sec:NTK}
The predictions and statistics of an infinite-width MLP of depth $L$ are fully characterized by two matrices: the kernel $K^{(L)}_{\alpha \beta}$ and the NTK $\Theta^{(L)}_{\alpha \beta}$, where $\alpha$ and $\beta$ are inputs from the training or test sets. In the notation of Ref.~\cite{Roberts:2021fes}, the components of these matrices can be defined through forward recursions
\begin{align}
\label{eq:Krec}
K^{(1)}_{\alpha \beta} = \frac{1}{n_0} \vec{x}_\alpha \cdot \vec{x}_\beta, \quad K_{\alpha \beta}^{(\ell+1)} & = \langle \sigma(u_\alpha) \sigma(u_\beta) \rangle_{K^{(\ell)}};  \\
\label{eq:NTKrec}
\Theta^{(1)}_{\alpha \beta} = \lambda_b + \frac{\lambda_W}{n_0} \vec{x}_\alpha \cdot \vec{x}_\beta; \quad \Theta_{\alpha \beta}^{(\ell+1)} & = \frac{\lambda_b}{\ell} + \lambda_W \left\langle \sigma(u_\alpha) \sigma(u_\beta) \right\rangle_{K^{(\ell)}}  +\left\langle \sigma'(u_\alpha) \sigma'(u_\beta) \right\rangle_{K^{(\ell)}} \Theta_{\alpha \beta}^{(\ell)},
\end{align}
where 
\begin{equation}
\langle F(u_{\alpha_1}, u_{\alpha_2}) \rangle_{K} \equiv  \frac{1}{\sqrt{{\rm det}(2 \pi \bK_2)}} \int du_{\alpha_1} \, du_{\alpha_2} \exp\left(-\frac{1}{2} \mathbf{u}^{\rm T} \bK{_2} \mathbf{u}\right) F(u_{\alpha_1}, u_{\alpha_2}) 
\end{equation}
is a two-dimensional Gaussian expectation over $\mathbf{u} \equiv (u_{\alpha_1}, u_{\alpha_2})$ with covariance $\bK_2 \equiv K_{\alpha_1 \alpha_2}$, the $2 \times 2$ submatrix of $K$. We choose $\sigma(z) = \erf(z)$ in this work in order to analytically evaluate all Gaussian integrals in~(\ref{eq:Krec}) and~(\ref{eq:NTKrec}), as derived in Ref.~\cite{williams1996computing,PANG2019270,lee2019wide}, which minimizes numerical error.

For notational convenience, we now specialize to the case of one output neuron i.e. $n_L = 1$, though all of the steps below generalize to any $n_L$. At initialization, the output distribution $p(z^{(L)} | \mathcal{A})$ is Gaussian with mean zero and covariance $K^{(L)}$. Each subsequent GD step is a linear transformation,
\begin{equation}
\label{eq:GDstep}
z^{(L)}_\delta \leftarrow z^{(L)}_\delta - \eta \sum_{\alpha \in \mathcal{A}} \Theta_{\delta \alpha}^{(L)} (z^{(L)}_\alpha - y_\alpha),
\end{equation}
which updates the mean and covariance of $p(z^{(L)} | \mathcal{A})$ but preserves the Gaussian nature of the distribution. If the NTK submatrix evaluated on the training set, $\bNTK_{\mathcal{A}}$, is invertible, taking the limit of an infinite number of GD steps leads to a convergent geometric series. In this case, the end-of-training mean prediction $m_{\beta}^{\infty}$ and variance $(\sigma_{\beta}^{\infty})^2$ for a single test point $\vec{x}_\beta$ is independent of $\eta$ and is given by~\cite{lee2019wide,Roberts:2021fes}:
\begin{equation}
\label{eq:infwidthpredmu}
m_{\beta}^{\infty} \equiv \mathbb{E} [ z_\beta^{(L)}] = \bNTK_\beta^{\rm T} {\bNTK}_{\mathcal{A}}^{-1} \mathbf{y}_{\mathcal{A}}, 
\end{equation}
\begin{equation}
\label{eq:infwidthpredsigma}
    \  (\sigma_{\beta}^{\infty})^2 \equiv {\rm Var}[z_\beta^{(L)}] = K^{(L)}_{\beta \beta} - 2 \bNTK_\beta^{\rm T} {\bNTK}_{\mathcal{A}}^{-1} \bK_{\beta} + \bNTK_\beta^{\rm T} {\bNTK}_{\mathcal{A}}^{-1} \mathbf{K}_{\mathcal{A}} {\bNTK}_{\mathcal{A}}^{-1} \bNTK_\beta,
\end{equation}
where $\mathbf{y}_{\mathcal{A}}$ is the vector of training set labels; $\mathbf{K}_{\mathcal{A}}$ is the kernel evaluated on the training set; and $\bNTK_\beta$ and $\bK_{\beta}$ are columns of $\Theta_{\alpha \beta}^{(L)}$ and $K_{\alpha \beta}^{(L)}$, respectively. If $\bNTK_{\mathcal{A}}$ is ill-conditioned, we may still compute the mean and variance of the prediction at each step of GD using Eq.~(\ref{eq:GDstep}), and implement early stopping to determine $m_{\beta}^{\infty}$ and $\sigma_{\beta}^{\infty}$. In either case, our desired quantities $\muL$ and $\sigmaL^2$ are
 \begin{equation}
 \label{eq:musigmaL}
 \muL \equiv \frac{1}{2|\mathcal{B}|} \sum_{\beta \in \mathcal{B}} \mathbb{E}\left [(z_{\beta}^{(L)} - y_{\beta})^2 \right]
\end{equation}
\begin{equation}
     \ \ \sigmaL^2 \equiv \left (\frac{1}{4|\mathcal{B}|^2} \sum_{\beta_1, \beta_2 \in \mathcal{B}} \mathbb{E} \left [ (z_{\beta_1}^{(L)} - y_{\beta_1})^2 (z_{\beta_2}^{(L)} - y_{\beta_2})^2\right]\right) - \muL^2.
\end{equation}
The Gaussian expectations are evaluated in Appendix~\ref{app:derivations} and generalized for $n_L \neq 1$.




\section{Finite-width Experiments}
\label{sec:experiments}

We compare the infinite-width predictions of Sec.~\ref{sec:NTK} with experiments on finite-width networks. We use three example datasets: the standard MNIST handwritten digit benchmark, the standard CIFAR color image classification benchmark, and an example relevant for HEP, namely simulated calorimeter data generated for a prior study of electron and photon classification and energy measurement using regression~\cite{Belayneh_2020}. A common application of machine learning in HEP is to determine the energy of a high-energy particle based on the pattern of energy deposits (``hits'') in a segmented calorimeter detector. Because the spatial distribution of hits is important, an architecture encoding spatial correlations such as a convolutional neural network (CNN) is likely preferable to an MLP, but a persistent issue is quantifying uncertainties in the inferred energy from these analyses; currently, UQ is done by training large ensembles of neural networks, which is resource-intensive \cite{Belayneh_2020}. To apply our MLP setup to this problem, we flatten the raw simulated calorimeter hits into a single vector (with the expectation of dramatic performance reduction) and regress to the true energy of the particle initiating the event. Details of the data generation procedure, as well as a link to our training set data, are given in Appendix~\ref{app:calo}. In the calorimeter example, $\muL \neq 0$ represents the magnitude of the bias away from the true energy of each event. If $\relvar \ll 1$, then $\muL$ should be assigned as the systematic uncertainty on the energy calibration. However, if $\relvar \gtrsim 1$, the fluctuations of the predictions of the MLP ensemble are larger than the bias, and the systematic uncertainty is instead dominated by $\sigmaL$.

\subsection{Training and architecture details}
In our experiments, we choose a random sample of size $\ND$ from each of our example datasets as our training set, and train an ensemble 150 MLPs for each $\ND$. Each MLP has $L = 3$ layers with a width of $n = 30$ for each hidden layer. The networks are initialized and trained as described in Sec.~\ref{sec:problem}. Training is executed using \texttt{TensorFlow}~\cite{tensorflow2015-whitepaper}, leveraging the NSF \texttt{ACCESS} program under an Explore allocation of $\sim$4000 node-hours. We set $\lambda_b/\lambda_W = 10$ and use $ \eta = 10$ for MNIST, $\eta = 5$ for CIFAR, and $\eta = 10^{-2}$ for the calorimeter dataset. For the calorimeter data at infinite width, we take $\eta = 10^{-4}$. Early stopping with a validation set of $10^3$ and patience of $10^4$ epochs is used to monitor for overfitting; each network trains in roughly 1 hour in $\approx 10^5-10^6$ epochs. Finally, the performance of each network is evaluated on a fixed test set of size $10^4$, which is consistent across all training set sizes. The resulting test loss of each network at each $\ND$ is saved to compute the sample mean and variance offline.

\begin{figure}[t!]
    \centering
    \begin{subfigure}{0.7\textwidth}
        \centering
        \includegraphics[width=\textwidth]{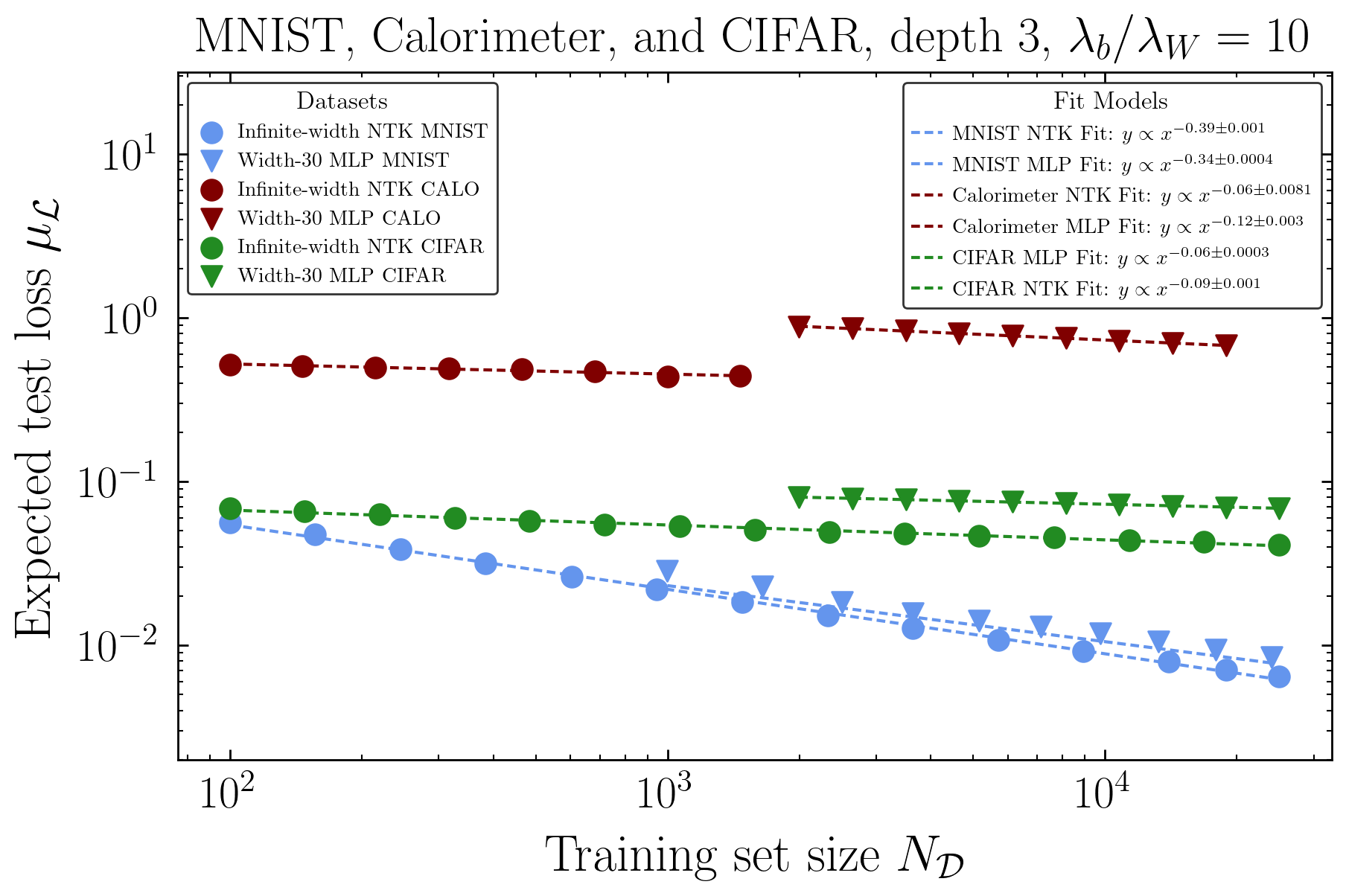}   
    \end{subfigure}
        \centering
    \begin{subfigure}{0.7\textwidth}
        \centering
     \includegraphics[width=\textwidth]{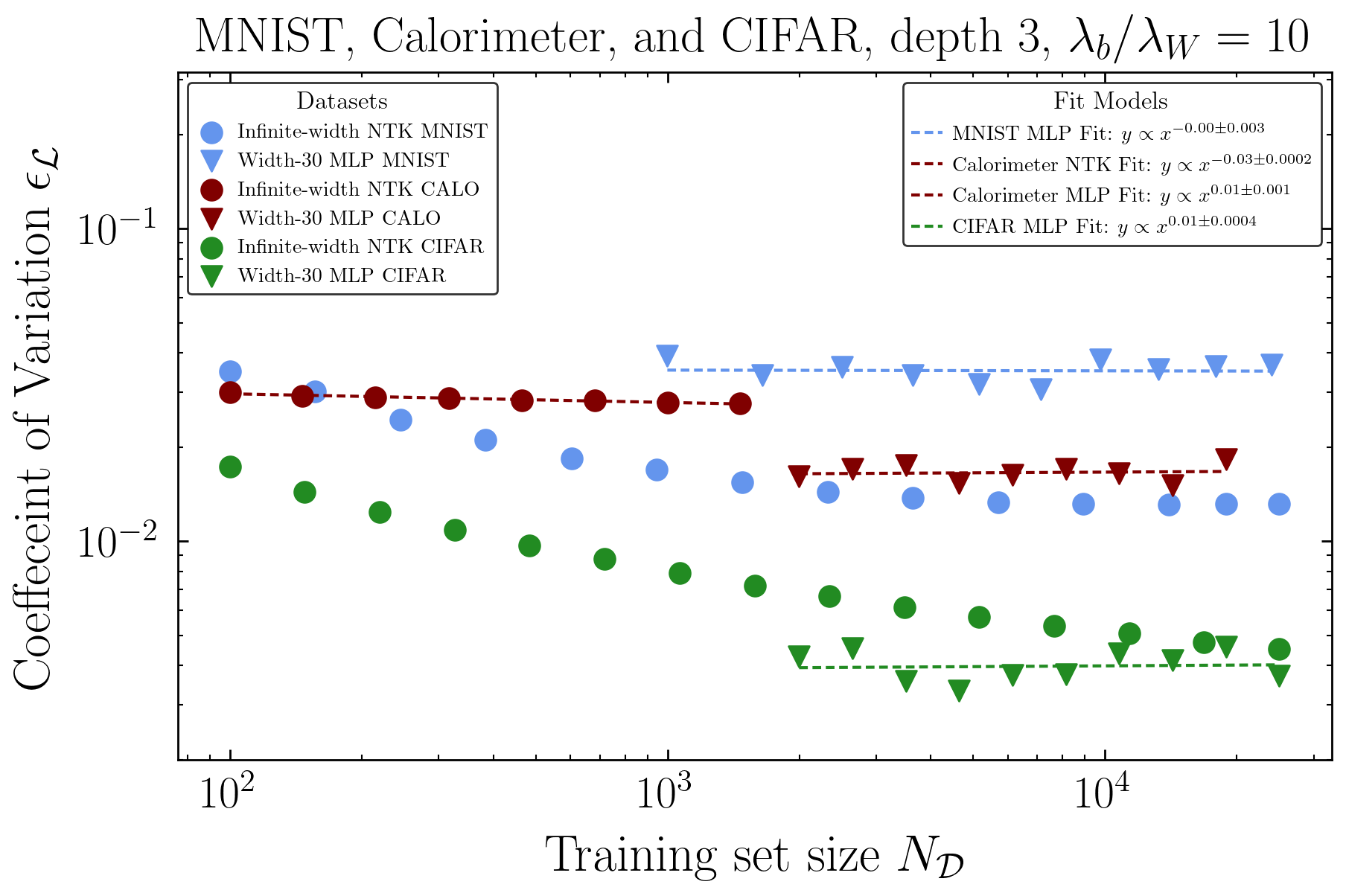}   
    \end{subfigure}
    \caption{Mean loss (top) and coefficient of variation (bottom) scaling laws on our three example datasets. The fits show linear least-squares fits in log-log space along with $1\sigma$ errors on the power law index. Infinite-width MNIST and CIFAR appear to show a broken power law for $\relvar$ so we do not provide a fit.}
    \vspace{-0.5cm}
    \label{fig:results}
\end{figure}

\subsection{Results}
The results of our experiments are shown in Fig.~\ref{fig:results}. We see that both the scaling and magnitude of the mean loss $\mu_{\mathcal{L}}$ (top panel) and coefficient of variation $\relvar$ (bottom panel) are quite similar between infinite and finite width. For our problems, $\muL$ does not differ by much for the two architectures, and similarly, $\relvar$ is on the same order of magnitude. Further, $\alpha_D \approx 0$ for $\relvar$ for infinite width and finite width for both datasets (or asymptotes to zero at large $\ND$ for infinite-width MNIST and CIFAR), highlighting $\relvar$ as a potential MLP ``invariant'' meriting future study. We see similar results for the \texttt{Adam} optimizer~\cite{kingma2014adam} (Appendix~\ref{app:adam}).

\subsection{A Plausibility Argument}
One can make the following heuristic plausibility argument for the infinite-width scaling of $\relvar$. Suppose the entries of the test-train NTK $\bNTK_{\mathcal{B}}$, kernel $\mathbf{K}$, and inverse NTK $\bNTK_{\mathcal{A}}^{-1}$ scale with $\ND$ as:
\begin{equation}
    \bNTK_{\mathcal{B}} \sim (\ND)^p,\quad \mathbf{K} \sim (\ND)^r,\quad \bNTK_{\mathcal{A}}^{-1} \sim (\ND)^k
\end{equation}
Empirically measuring each of these scaling laws (Appendix~\ref{app:KNscaling}), we find that $p, r \approx0$, but $k<0$. Thus the $\ND$ scaling of the inverse NTK elements seems to be driving the $\muL$ scaling law.  Using Eqs.~(\ref{eq:infwidthpredmu})--(\ref{eq:infwidthpredsigma}) and assuming the last term in $(\sigma_{\beta}^{\infty})^2$ dominates, we find $\muL \sim (\ND)^{2k+4}$ and $\sigmaL^2 \sim (\ND)^{4k+8}$, where the extra powers of $\ND$ come from the matrix multiplications over the training set. Since $\sigmaL^2$ contains exactly twice as many powers of $\bNTK_{\mathcal{A}}^{-1}$ and sums over $\mathcal{A}$ as $\muL$, the coefficient of variation scales as $\relvar \sim \frac{\sqrt{(\ND^{4k+8})}}{(\ND)^{2k+4}}\sim (\ND)^{0}$. The broken power law in $\relvar$ for MNIST and CIFAR may result from the different scaling of other terms in Eq.~(\ref{eq:infwidthpredmu}) for small $\ND$. To connect with existing UQ approaches in physics using Bayesian networks~\cite{Bollweg:2019skg,Kasieczka:2020vlh}, we investigate a Bayesian interpretation of the infinite-width $\relvar$ in Appendix~\ref{app:Bayesian}.

\subsection{Dependence on Learning Rate Ratio}
\label{sec:LearningRateRatio}
\begin{figure}[t!]
    \centering
    \begin{subfigure}{0.7\textwidth}
        \centering
        \includegraphics[width=\textwidth]{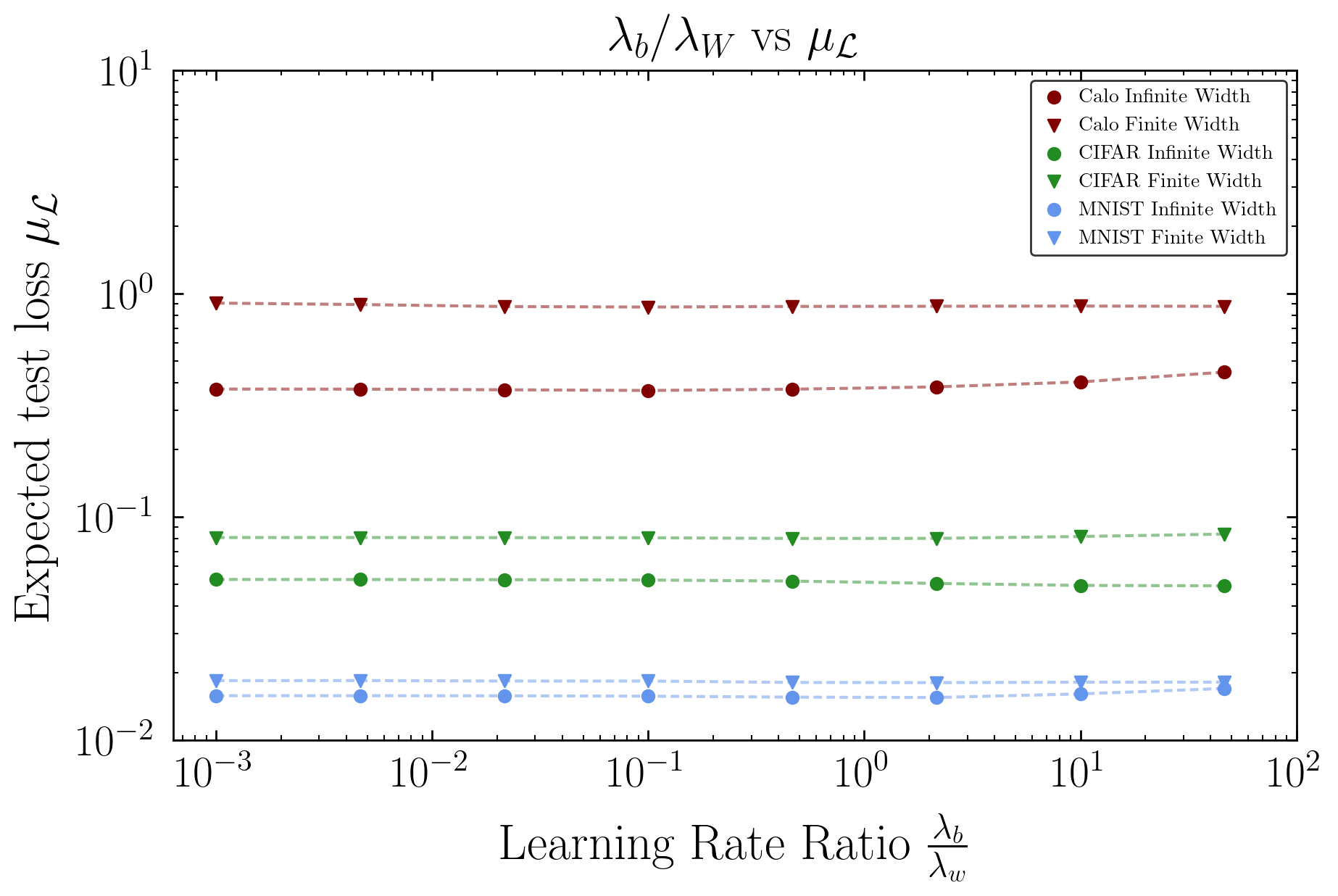}   
    \end{subfigure}
        \centering
    \begin{subfigure}{0.7\textwidth}
        \centering
     \includegraphics[width=\textwidth]{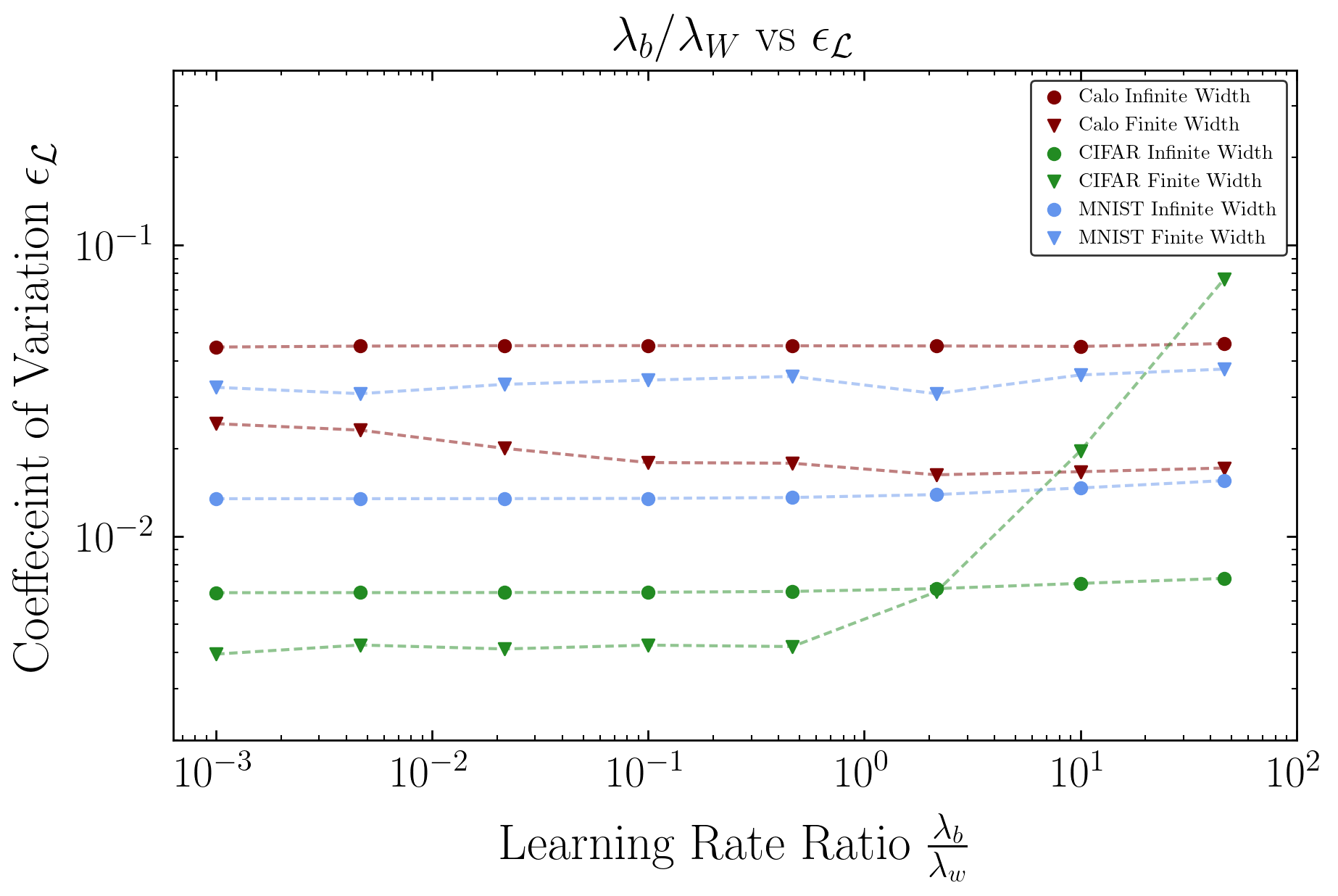}   
    \end{subfigure}
    \caption{Dependence on the learning rate ratio $\lambda_b/\lambda_W$ for mean loss (top) and coefficient of variation (bottom), on our three example datasets. For all datasets the dependence of $\mu_{\mathcal{L}}$ on $\lambda_b/\lambda_W$ is very weak, and the $\relvar$ dependence indicates a mild preference for $\lambda_b/\lambda_W = \mathcal{O}(1)$.}
    \vspace{-0.5cm}
    \label{fig:results_lambda}
\end{figure}
With fixed architecture and an invertible NTK, the only hyperparameter governing the infinite-width predictions is the ratio of bias to weight learning rate, $\lambda_b/\lambda_W$. We repeat the experiments of Sec.~\ref{sec:experiments}, fixing the train set size to $\ND=2000$, the weight learning rate to $\lambda_W=1$, and varying $\lambda_b$ from $10^{-3}$ to $10^3$. To guarantee convergence, we dynamically vary $\eta$ in our finite width experiments by setting $\eta \to \frac{\eta}{\lambda_b}$ for $\lambda_b>1$. We show the results in Fig.~\ref{fig:results_lambda}; it is clear that the performance at both infinite and finite width, as measured by $\mu_{\mathcal{L}}$, has very mild dependence on this hyperparameter. The same is true for the coefficient of variation $\relvar$, with the exception of CIFAR, where we see a sharp rise in the coefficient of variation at finite width for $\lambda_b/\lambda_W > 1$. With the exception of this curious feature, we see agreement at the $\mathcal{O}(1)$ between the infinite width predictions and finite width experiments for both $\mu_{\mathcal{L}}$ and $\relvar$ as a function of $\lambda_b/\lambda_W$, suggesting that a hyperparameter tuning to minimize $\relvar$ at infinite width can generalize to finite width.

\section{Conclusion}
In this work we have shown empirically that the mean and variance of the MLP test loss on three regression tasks exhibit scaling laws with training set size $\ND$. The scaling exponents for the mean loss $\muL$ resemble the corresponding scaling laws at infinite width, while the scaling exponent for the coefficient of variation $\relvar$ is approximately zero at finite width and asymptotes to zero at infinite width for large $\ND$. These results already indicate the importance of finite-width effects for UQ, which can change the scaling laws for $\sigmaL$ by order-1 factors. However, $\relvar$ may be well-approximated by the infinite-width value implying that systematic uncertainty is dominated by $\muL$ alone. It is interesting to note that in the NTK parameterization, feature learning is absent at infinite width but present finite width~\cite{Roberts:2021fes}, which suggests that a prediction of $\relvar$ may not rely on feature learning, unlike the prediction for the loss~\cite{vyas2023empirical,bordelon2024featurelearningimproveneural}.\footnote{That said, it may be the case that the overall size of $\relvar$ does rely on feature learning and/or the training algorithm; see App.~\ref{app:adam}.} We have also shown that $\relvar$ is largely independent of the learning rate ratio $\lambda_b/\lambda_W$ at both infinite width and finite width. 

In future work, it would be interesting to understand how our results relate to existing work using $\mu$-parametrization and compute $\relvar$ at finite width by generalizing the results of Sec.~\ref{sec:NTK} using finite-width perturbation theory. Indeed, the latter may be feasible even in light of the $\ND^4$ scaling of the finite-width tensors because of the independence of $\relvar$ from $\ND$, as long as the scaling laws persist to sufficiently small $\ND$. Finally, it would be interesting to apply finite-width perturbation theory to Bayesian neural networks~\cite{pacelli2023statistical} to see if the same scaling laws appear. 

\section*{Acknowledgments}

We thank Tilman Plehn, Christoph Weniger, and the participants of the IAIFI 2024 Summer Workshop, as well as Boris Hanin and Cengiz Pehlevan, for stimulating conversations. We also thank Hassan Sajjad, Becky Nevin, and the organizers and attendees of the FAIR Universe Competition at NeurIPS 2024 for useful insights. This material is based upon work supported by the U.S. Department of Energy, Office of Science, Office of High Energy Physics, under Award Number DE-SC0023704. This work utilizes the computing resources of the HAL cluster~\cite{10.1145/3311790.3396649} supported by the National Science Foundation’s Major Research Instrumentation program, grant \#1725729, as well as the University of Illinois Urbana-Champaign. This work used the \texttt{TAMU FASTER} cluster at Texas A\&M University through allocation 240449 from the Advanced Cyberinfrastructure Coordination Ecosystem: Services \& Support (ACCESS) program~\cite{10.1145/3569951.3597559}, which is supported by National Science Foundation grants \#2138259, \#2138286, \#2138307, \#2137603, and \#213829.

\appendix

\section{Derivation of mean loss and loss variance at infinite width}
\label{app:derivations}
Here we evaluate the Gaussian expectations to compute the mean test loss and its variance for infinite-width neural networks, given their mean predictions $m^\infty_{\beta}$ and the covariance matrix $\Sigma_{\beta_1 \beta_2}$. 

We start with the case where $n_L = 1$. The diagonal entries $\Sigma_{\beta \beta} \equiv (\sigma_\beta^\infty)^2$ were given in Eq.~(\ref{eq:infwidthpredsigma}); the full covariance matrix entries are~\cite{lee2019wide}
\begin{equation}
\Sigma_{\beta_1 \beta_2} = K^{(L)}_{\beta_1 \beta_2} - \bNTK_{\beta_1}^{\rm T} {\bNTK}_{\mathcal{A}}^{-1} \bK_{\beta_2} - \bNTK_{\beta_2}^{\rm T} {\bNTK}_{\mathcal{A}}^{-1} \bK_{\beta_1} + \bNTK_{\beta_1}^{\rm T} {\bNTK}_{\mathcal{A}}^{-1} \mathbf{K}_{\mathcal{A}} {\bNTK}_{\mathcal{A}}^{-1} \bNTK_{\beta_2}.
\end{equation}
It will be convenient to shift the output $z^{(L)}_{\beta}$ by the mean, so that $\widetilde{z}^{(L)}_\beta \equiv z^{(L)}_{\beta} - m_\beta^\infty$ are zero-mean Gaussian variables with the same covariance $\Sigma_{\beta_1 \beta_2}$ as the unshifted outputs. Defining $\Delta_\beta \equiv  y_\beta^\infty - m_\beta$ as the mean prediction error for each test point $\vec{x}_\beta$, we have~\cite{Roberts:2021fes}
\begin{align}
\muL \equiv \mathbb{E}[\Loss_B(T)] & = \mathbb{E}\left[\frac{1}{2|\mathcal{B}|}  \sum_{\beta \in B} (z^{(L)}_{\beta} - y_{\beta})^2\right] \nonumber \\
&= \frac{1}{2|\mathcal{B}|} \sum_{\beta \in B}  \mathbb{E}\left[ (\zt_\beta - \Delta_\beta )^2\right] \nonumber \\
&= \frac{1}{2|\mathcal{B}|}  \sum_{\beta \in B} \left( \Delta_\beta^2 + \Sigma_{\beta \beta} \right),
\label{eq:muL}
\end{align}
where in the last line we have used $\mathbb{E} \left [ \zt_\beta \right] = 0$ and $\mathbb{E} \left [ (\zt_\beta)^2 \right] = \Sigma_{\beta \beta}$. 

For the variance, we additionally need to compute
\begin{align}
\mathbb{E}[\Loss_{\mathcal{B}}^2] & = \frac{1}{4|\mathcal{B}|^2} \sum_{\beta_1, \beta_2 \in \mathcal{B}} \mathbb{E} \left [ (z_{\beta_1}^{(L)} - y_{\beta_1})^2 (z_{\beta_2}^{(L)} - y_{\beta_2})^2\right] \nonumber \\
& = \frac{1}{4|\mathcal{B}|^2} \sum_{\beta_1, \beta_2 \in \mathcal{B}} \left \{ \mathbb{E} \left [ (\zt_{\beta_1})^2 (\zt_{\beta_2})^2 \right ] + \Delta_{\beta_1}^2 \mathbb{E} \left [ (\zt_{\beta_2})^2 \right]  + \Delta_{\beta_2}^2 \mathbb{E} \left [ (\zt_{\beta_1})^2 \right] \right. \nonumber \\
& \left. \hspace{2.5cm} + \,  4 \Delta_{\beta_1} \Delta_{\beta_2} \mathbb{E} \left [ \zt_{\beta_1} \zt_{\beta_2} \right] + \Delta_{\beta_1}^2 \Delta_{\beta_2}^2 \right \} \nonumber \\
& \hspace{-1.2cm} = \frac{1}{4|\mathcal{B}|^2} \sum_{\beta_1, \beta_2 \in \mathcal{B}} \left \{ \Sigma_{\beta_1 \beta_1}\Sigma_{\beta_2 \beta_2} + 2 \Sigma_{\beta_1 \beta_2}^2 + \Delta_{\beta_1}^2 \Sigma_{\beta_2 \beta_2} + \Delta_{\beta_2}^2 \Sigma_{\beta_1 \beta_1} + 4 \Delta_{\beta_1} \Delta_{\beta_2} \Sigma_{\beta_1 \beta_2} + \Delta_{\beta_1}^2 \Delta_{\beta_2}^2 \right \},
\label{eq:LBsq}
\end{align}
where in the second line we have used the fact that odd moments of $\zt_\beta$ vanish, and in the third line we have used Wick's theorem for the quartic expectation. The loss variance is then given by
\begin{equation}
\sigmaL^2 = \mathbb{E}[\Loss_{\mathcal{B}}^2] - \muL^2,
\end{equation}
where the first term is given in Eq.~(\ref{eq:LBsq}) and the second term is given in Eq.~(\ref{eq:muL}).

For $n_L \neq 1$, we index the output neurons with Latin letters $i, j$: $\widetilde{z}^{(L)}_{\beta,i} \equiv z^{(L)}_{\beta,i} - m_{\beta,i}^\infty$ and $\Delta_{\beta,i} \equiv  y_{\beta,i}^\infty - m_{\beta,i}$. At infinite width, output neurons are perfectly uncorrelated so the covariance matrix is diagonal in neural indices,
\begin{equation}
\mathbb{E} \left [ \zt_{\beta_1,i} \zt_{\beta_1,j}\right] = \delta_{ij}\Sigma_{\beta_1 \beta_2}.
\end{equation}
The mean loss is
\begin{align}
\muL & = \mathbb{E}\left[\frac{1}{2|\mathcal{B}|n_L}  \sum_{\beta \in B} \sum_{i = 1}^{n_L} (z^{(L)}_{\beta,i} - y_{\beta,i})^2\right] \nonumber \\
&= \frac{1}{2|\mathcal{B}|n_L} \sum_{\beta \in B} \sum_{i = 1}^{n_L} \mathbb{E}\left[ (\zt_{\beta_i} - \Delta_{\beta,i} )^2\right] \nonumber \\
&= \frac{1}{2|\mathcal{B}|n_L}  \sum_{\beta \in B} \left ( n_L \Sigma_{\beta \beta} +\sum_{i = 1}^{n_L}  \Delta_{\beta,i}^2 \right) \nonumber \\
& = \frac{1}{2|\mathcal{B}|n_L}  \sum_{\beta \in B} \left ( n_L \Sigma_{\beta \beta} + ||\vec{\Delta}_{\beta}||^2 \right),
\label{eq:muLnL}
\end{align}
where the factor of $n_L$ comes from summing over $n_L$ identical copies of the covariance $\mathbb{E} \left [ (\zt_{\beta,i})^2 \right]$, and in the last line we have changed to vector notation, $\sum_{i = 1}^{n_L}  \Delta_{\beta,i}^2 \equiv ||\vec{\Delta}_{\beta}||^2 = \vec{\Delta}_{\beta} \cdot \vec{\Delta}_{\beta}$. 

Similarly, for the variance we need
\begin{align}
\mathbb{E}[\Loss_{\mathcal{B}}^2] & = \frac{1}{(2|\mathcal{B}|n_L)^2} \sum_{\beta_1, \beta_2 \in \mathcal{B}} \sum_{i,j = 1}^{n_L} \mathbb{E} \left [ (z_{\beta_1,i}^{(L)} - y_{\beta_1,i})^2 (z_{\beta_2,j}^{(L)} - y_{\beta_2,j})^2\right] \nonumber \\
& = \frac{1}{(2|\mathcal{B}|n_L)^2} \sum_{\beta_1, \beta_2 \in \mathcal{B}}  \sum_{i,j = 1}^{n_L} \left \{ \mathbb{E} \left [ (\zt_{\beta_{1,i}})^2 (\zt_{\beta_{2,j}})^2 \right ] + \Delta_{\beta_{1,i}}^2 \mathbb{E} \left [ (\zt_{\beta_{2,j}})^2 \right]  + \Delta_{\beta_{2,j}}^2 \mathbb{E} \left [ (\zt_{\beta_{1,i}})^2 \right] \right. \nonumber \\
& \left. \hspace{2.5cm} + \,  4 \Delta_{\beta_{1,i}} \Delta_{\beta_{2,j}} \mathbb{E} \left [ \zt_{\beta_{1,i}} \zt_{\beta_{2,j}} \right] + \Delta_{\beta_{1,i}}^2 \Delta_{\beta_{2,j}}^2 \right \} \nonumber \\
&  = \frac{1}{(2|\mathcal{B}|n_L)^2} \sum_{\beta_1, \beta_2 \in \mathcal{B}} \Bigg [ n_L^2 \Sigma_{\beta_1 \beta_1}\Sigma_{\beta_2 \beta_2} + 2 n_L \Sigma_{\beta_1 \beta_2}^2 + n_L \sum_{i = 1}^{n_L} \left \{ \Delta_{\beta_{1,i}}^2 \Sigma_{\beta_2 \beta_2} + \Delta_{\beta_{2,i}}^2 \Sigma_{\beta_1 \beta_1} \right \} \nonumber \\
& \hspace{2.5cm} + 4 \Sigma_{\beta_1 \beta_2} \sum_{i = 1}^{n_L} \{ \Delta_{\beta_{1,i}} \Delta_{\beta_{2,i}} \}  + \sum_{i,j = 1}^{n_L} \Delta_{\beta_{1,i}}^2 \Delta_{\beta_{2,j}}^2 \Bigg ] \nonumber \\
& = \frac{1}{(2|\mathcal{B}|n_L)^2} \sum_{\beta_1, \beta_2 \in \mathcal{B}} \Bigg [ n_L^2 \Sigma_{\beta_1 \beta_1}\Sigma_{\beta_2 \beta_2}  + n_L \left(2  \Sigma_{\beta_1 \beta_2}^2 + ||\vec{\Delta}_{\beta_1}||^2 \, \Sigma_{\beta_2 \beta_2} + ||\vec{\Delta}_{\beta_2}||^2 \, \Sigma_{\beta_1 \beta_1}  \right ) \nonumber \\
& \hspace{2.5cm} + 4 \Sigma_{\beta_1 \beta_2} \vec{\Delta}_{\beta_1} \cdot \vec{\Delta}_{\beta_2} + ||\vec{\Delta}_{\beta_1}||^2 \, ||\vec{\Delta}_{\beta_2}||^2\Bigg ].
\label{eq:LBsqnL}
\end{align}
As before, the variance $\sigmaL^2$ is computed by subtracting Eq.~(\ref{eq:muLnL}) from Eq.~(\ref{eq:LBsqnL}).

\section{Calorimeter Data Details}
\label{app:calo}
Particle colliders and their detectors are the prototypical tools in HEP. These detectors include calorimeters, which measure particles' energies and are composed of an array of detecting elements, or “cells,” where particles deposit energy. This results in a spatial distribution of energy deposition that is used to both identify entering particles and quantify their energies. Thus, machine learning can be used for both (particle) classification and (energy) regression; in this work we focus on the latter.

Following Ref.~\cite{Belayneh_2020}, we generate our dataset using \texttt{GEANT4}~\cite{GEANT4:2002zbu}, simulating the proposed Linear Collider Detector (LCD) at the CLIC accelerator. In the LCD, we simulate an electromagnetic calorimeter (ECAL) and a hadronic calorimeter (HCAL). The ECAL is structured with 25 silicon sensor planes interspersed with tungsten absorber layers and arranged in a cylindrical geometry with square cells measuring 5.1 mm on each side. The HCAL, positioned behind the ECAL, comprises 60 layers of polystyrene scintillators with steel absorbers, segmented into larger $3 \times 3 \ {\rm cm}^2$ square cells. The dataset is formed by outputting 3D arrays of energy deposits in each of the ECAL and HCAL cells, which due to their geometry gives us, for each event, an $11\times 11 \times 60$ array for HCAL and a $25 \times 25 \times 25$ array for ECAL. These arrays are then flattened into a large 22885-dimensional vector. 

While many particles can in principle be simulated, we specialize to a sample of $10^5$ electrons entering the HCAL and ECAL (where one electron entering corresponds to an event and thus a datapoint) with a range of incident angles close to the beam direction and a range of $10-100$ GeV of incoming energy. This incoming energy is set as the label corresponding to our 22885-dimensional input vector. We take into account the effects of solenoidal magnetic field and other materials preceding the calorimeters.

Our dataset of $10^5$ electron events can be found here: \url{https://zenodo.org/records/13715377}.

\section{Variation of training algorithm}
\label{app:adam}

\begin{figure}[htbp]
    \centering
    \begin{subfigure}{0.7\textwidth}
        \centering
     \includegraphics[width=\textwidth]{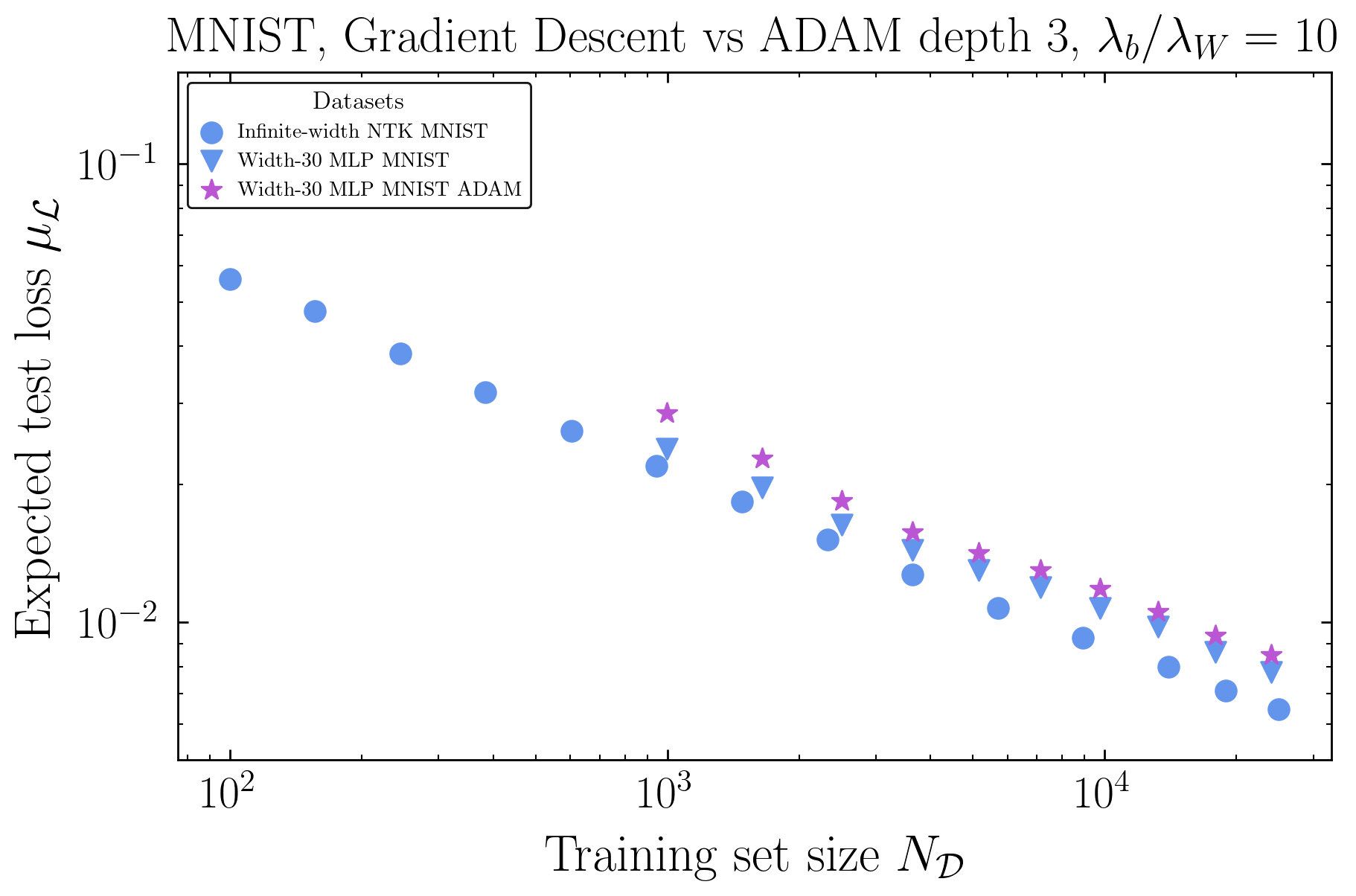}   
    \end{subfigure}\hfill
    \begin{subfigure}{0.7\textwidth}
        \centering
        \includegraphics[width=\textwidth]{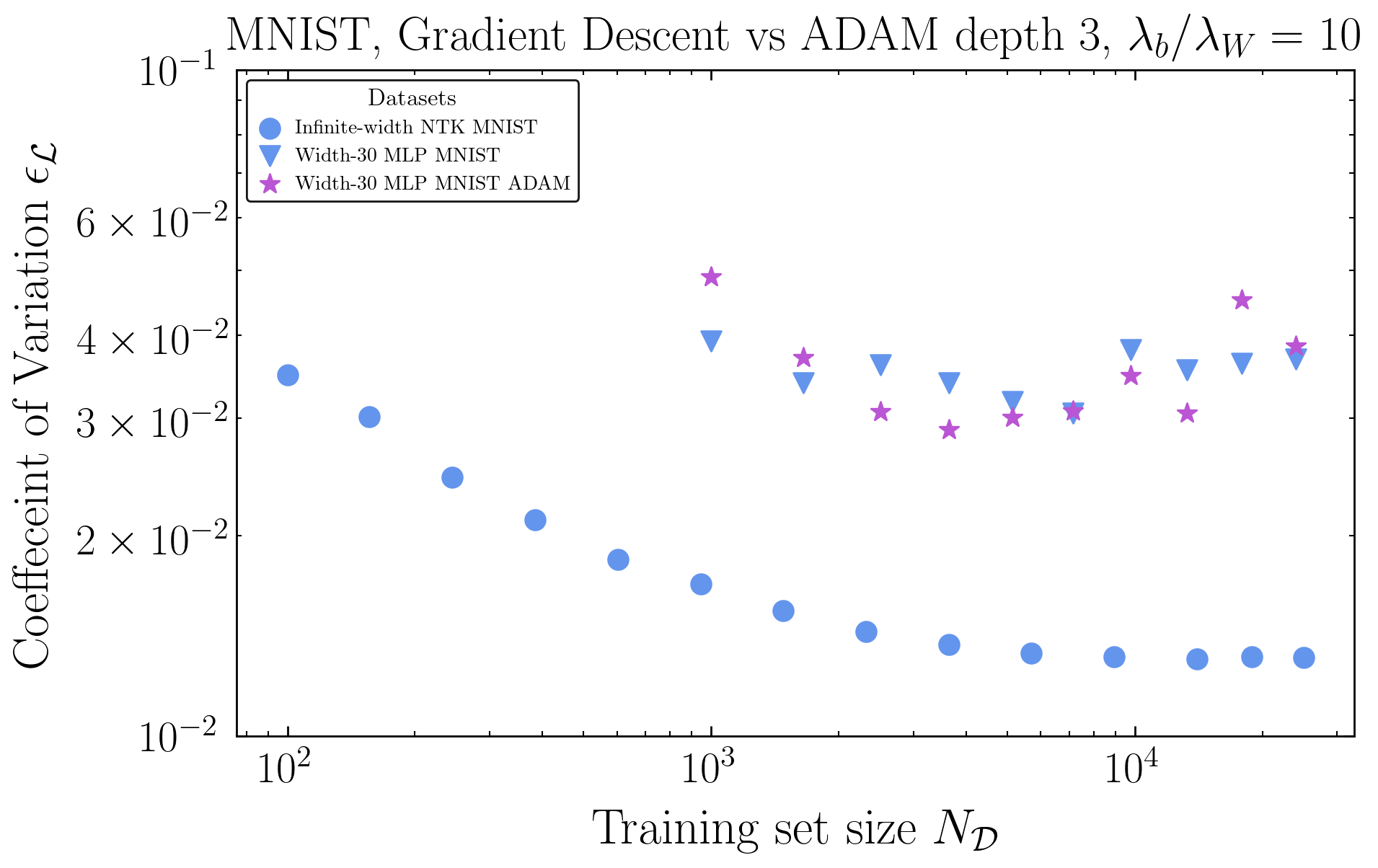}   
    \end{subfigure}
    \caption{Mean loss (top) and coefficient of variation (bottom) scaling laws using the \texttt{Adam} optimizer in MNIST. For this example, \texttt{Adam} and full-batch GD show very similar scaling laws for both $\muL$ and $\relvar$.}
    \label{fig:adamMNIST}
\end{figure}

\begin{figure}[htbp]
    \centering
    \begin{subfigure}{0.7\textwidth}
        \centering
     \includegraphics[width=\textwidth]{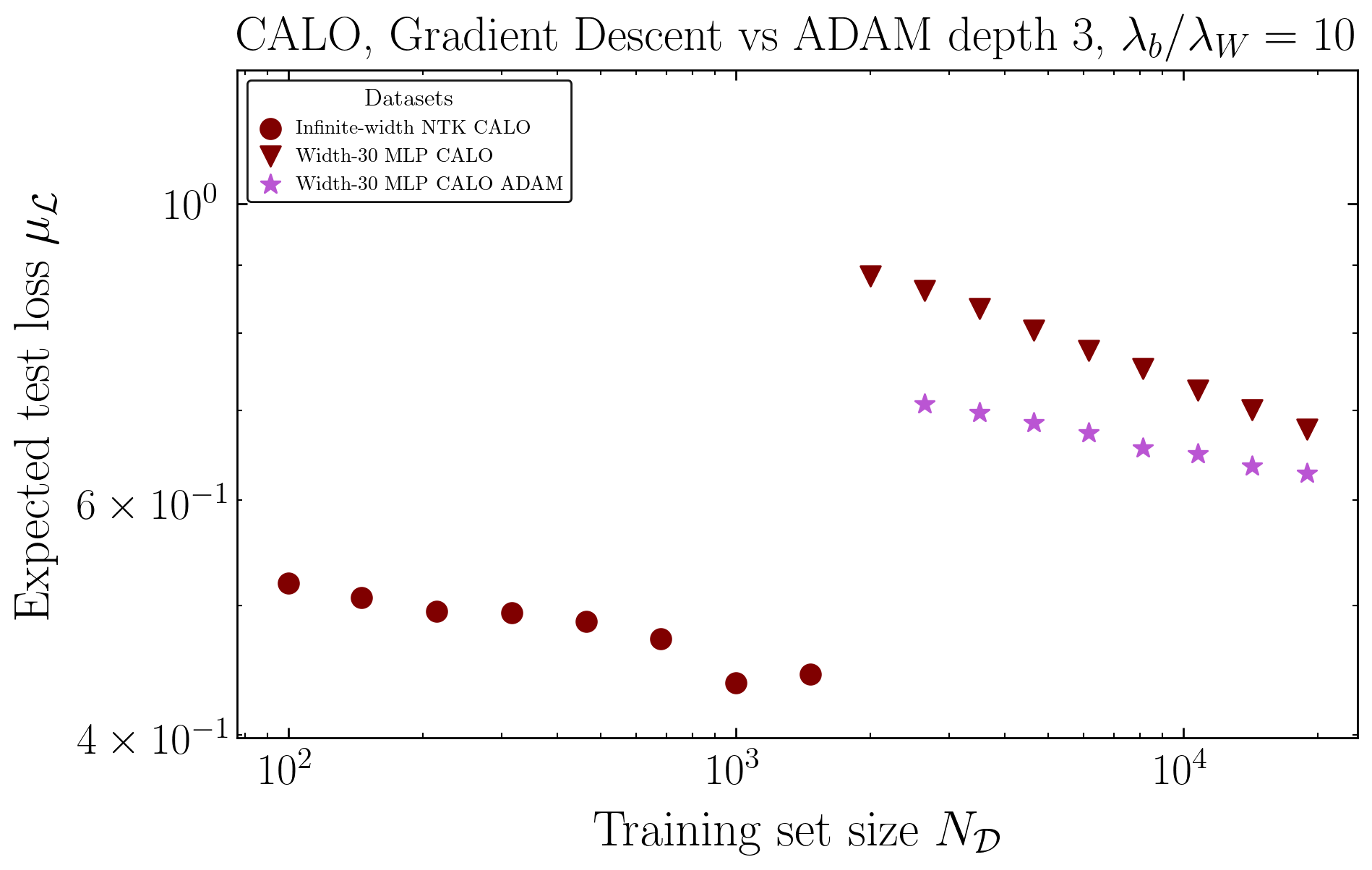}   
    \end{subfigure}\hfill
    \begin{subfigure}{0.7\textwidth}
        \centering
        \includegraphics[width=\textwidth]{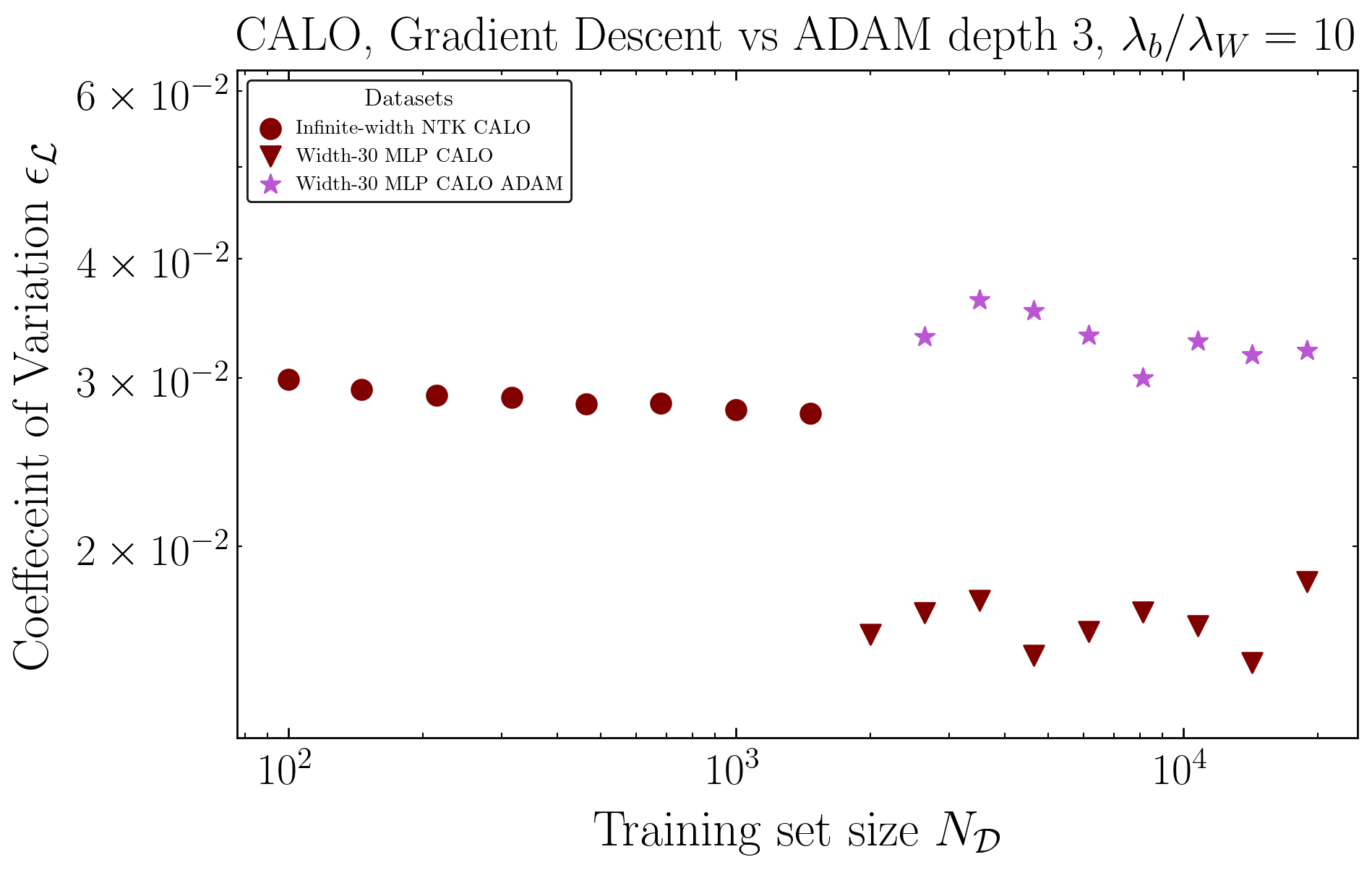}   
    \end{subfigure}
    \caption{Mean loss (top) and coefficient of variation (bottom) scaling laws using the \texttt{Adam} optimizer in calorimeter data. For this example, the $\relvar$ scaling appears to be flat for \texttt{Adam} as it was for full-batch GD, but the $\muL$ scaling law is markedly different.}
    \label{fig:adamCALO}
\end{figure}

While our infinite-width results apply, strictly speaking, only for full-batch gradient descent, we can investigate how robust the $\ND$ independence of $\relvar$ is when we vary the training algorithm. In Figs.~\ref{fig:adamMNIST} and \ref{fig:adamCALO} we show the mean test loss and coefficient of variation as in Fig.~\ref{fig:results}, but using networks trained with the \texttt{Adam} optimizer with $\eta=10^{-3}$ for MNIST and $\eta=10^{-6}$ for calorimter data. For both datasets, we use $\beta_1=.9$, $\beta_2 = .999$ and with a minibatch size of 1000. The results are somewhat noisier, but the overall trend is for a roughly flat scaling law for $\relvar$ (right panels). However, while the $\muL$ scaling laws for MNIST agree almost perfectly between the two optimizers, the $\muL$ scaling laws for the calorimeters are markedly different, and the values for $\relvar$ differ by more than a factor of 2. It would be interesting to investigate in future work to what extent the different behavior of MNIST and the calorimeter data is a consequence of finite-width feature learning.

\section{Measurements of kernel and NTK element scalings with dataset size}
\label{app:KNscaling}

\begin{figure}[htbp]
    \centering
    \begin{subfigure}{0.7\textwidth}
        \centering
     \includegraphics[width=\textwidth]{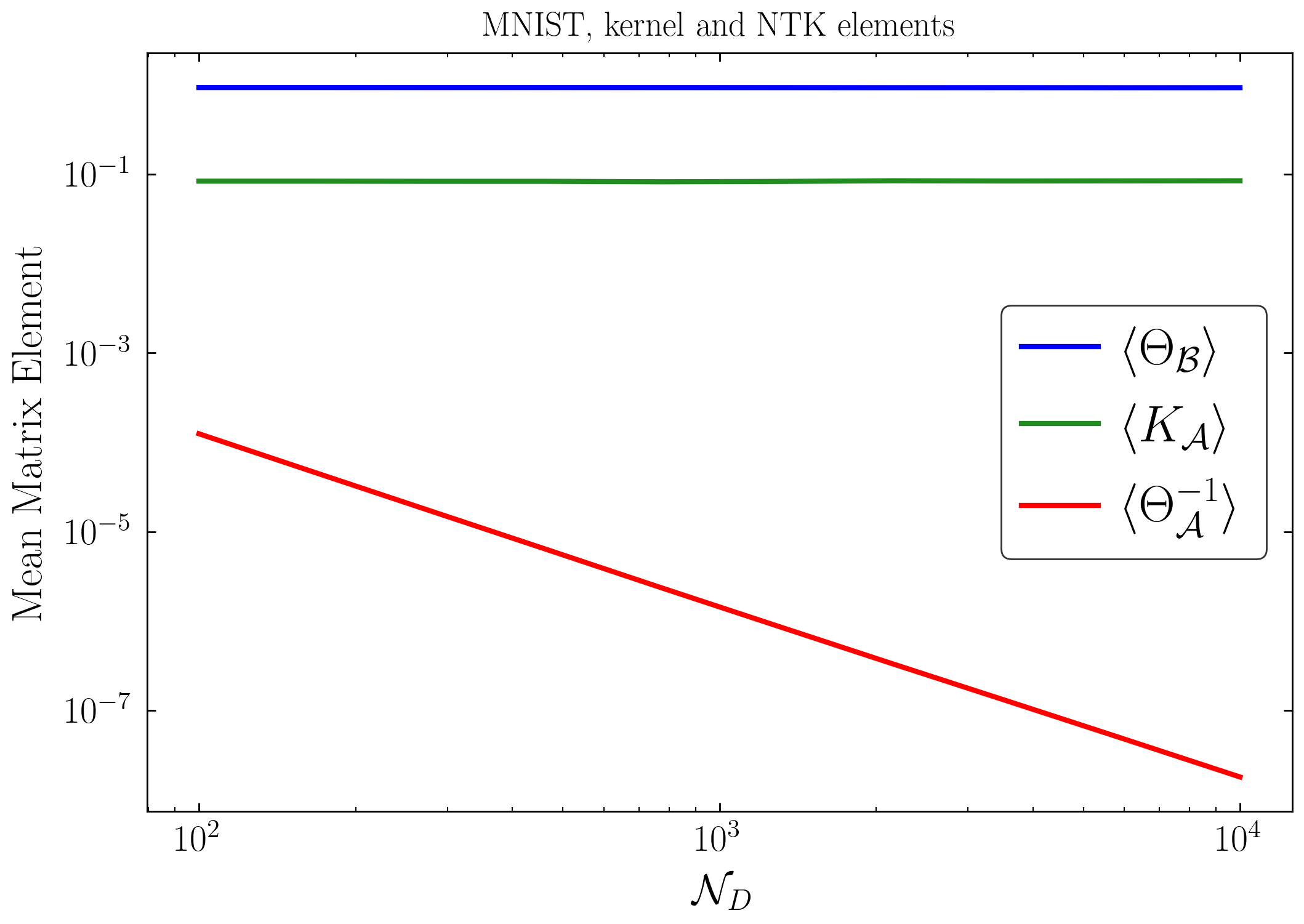}   
    \end{subfigure}\hfill
    \begin{subfigure}{0.7\textwidth}
        \centering
        \includegraphics[width=\textwidth]{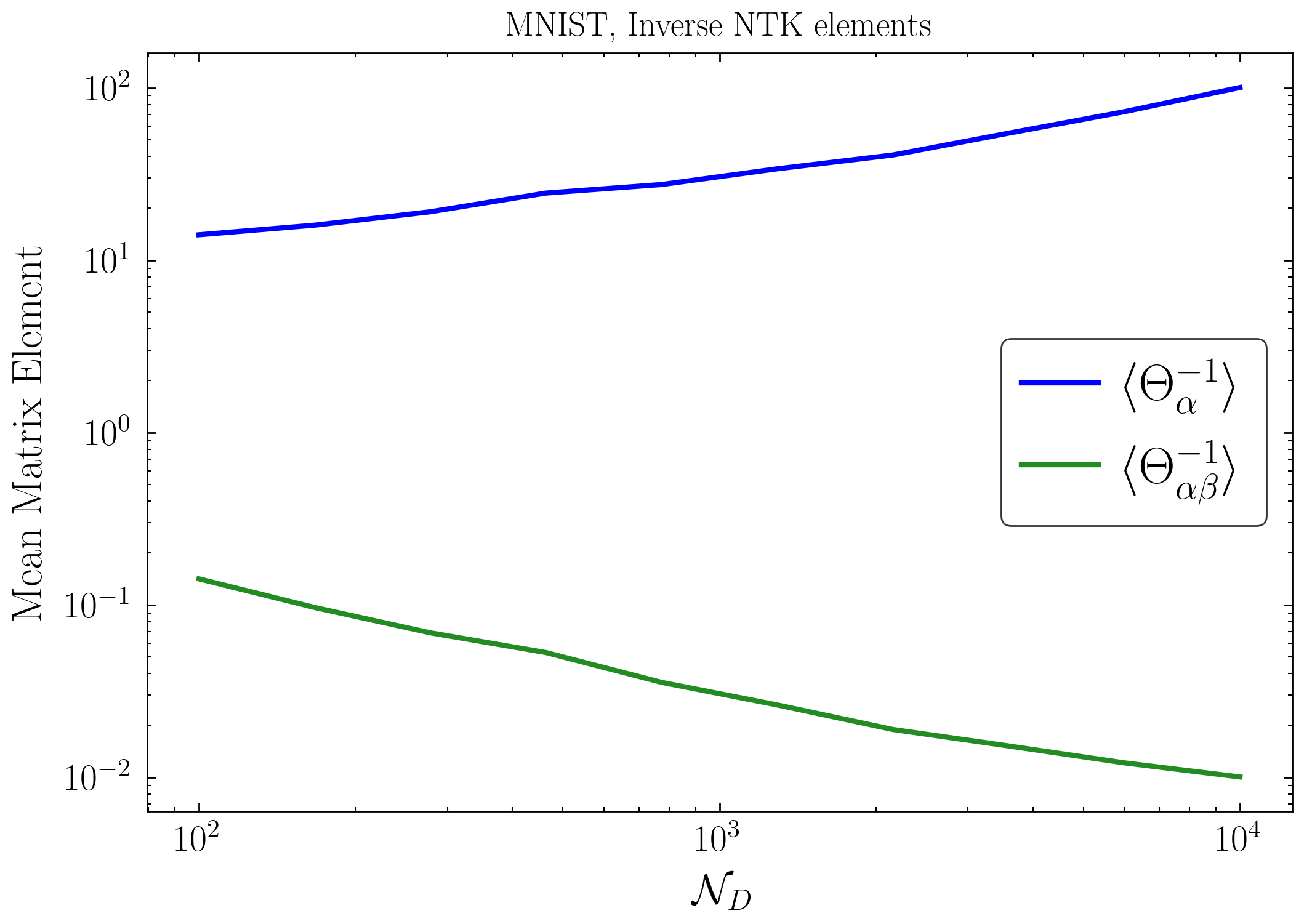}   
    \end{subfigure}
    \caption{Scaling of mean matrix elements with training set size $\ND$. Top: The only matrix contributing to the mean and variance of the infinite-width prediction that appears to scale nontrivially with $\ND$ is the training NTK inverse $\bNTK^{-1}_{\mathcal{A}}$. Bottom: the diagonal and off-diagonal elements of the NTK scale differently.}
    \label{fig:matscalings}
\end{figure}

For the case where the NTK is invertible, we can attempt to gain some intuition for the $\ND$ independence of the coefficient of variation statistic $\relvar$ by ignoring the matrix structure of the infinite-width predictions~(\ref{eq:infwidthpredmu})--(\ref{eq:infwidthpredsigma}) and simply studying the scaling behavior of the kernel and NTK elements. In our example datasets, the NTK is only numerically invertible for MNIST. In Fig.~\ref{fig:matscalings} we show the mean matrix elements for the MNIST test-train NTK $\bNTK_{\mathcal{B}}$, the training set kernel $\mathbf{K}_{\mathcal{A}}$, and the inverse of the training NTK, $\bNTK^{-1}_{\mathcal{A}}$. Only the latter scales nontrivially with $\ND$ (top panel). In the bottom panel, we show the scalings of the mean diagonal and off-diagonal elements, $\bNTK^{-1}_{\alpha \alpha}$ and $\bNTK^{-1}_{\alpha \beta}$, respectively, of the inverse NTK. The scalings are different, with the diagonal elements growing with $\ND$ while the off-diagonal elements decrease. It would be interesting to investigate further how these scaling laws relate to generalization behavior in the infinite-width limit.

\section{Bayesian interpretation of infinite-width ensemble}
\label{app:Bayesian}
\begin{figure}[htbp]
    \centering
    \begin{subfigure}{0.7\textwidth}
        \centering
     \includegraphics[width=\textwidth]{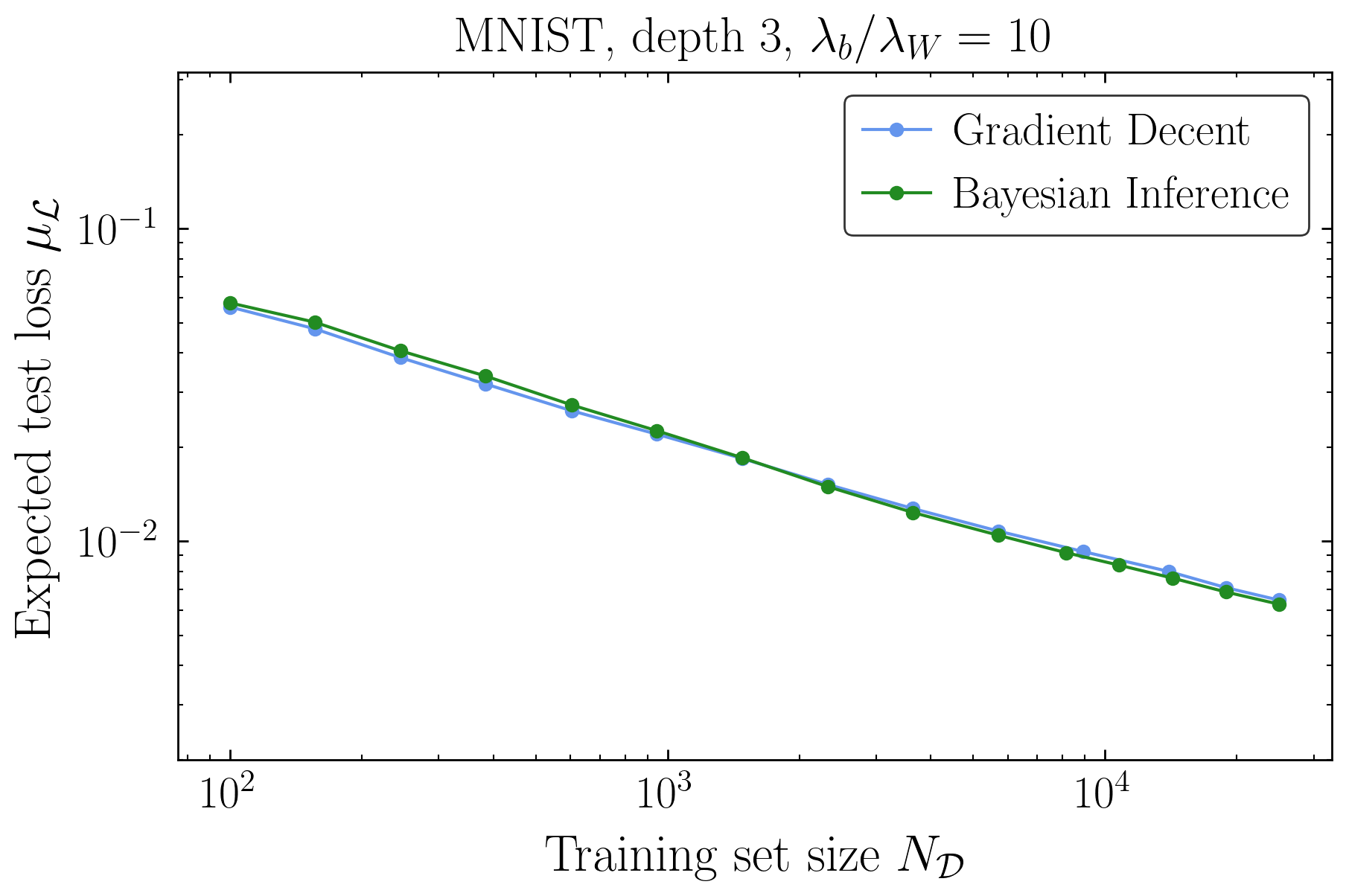}   
    \end{subfigure}\hfill
    \begin{subfigure}{0.7\textwidth}
        \centering
        \includegraphics[width=\textwidth]{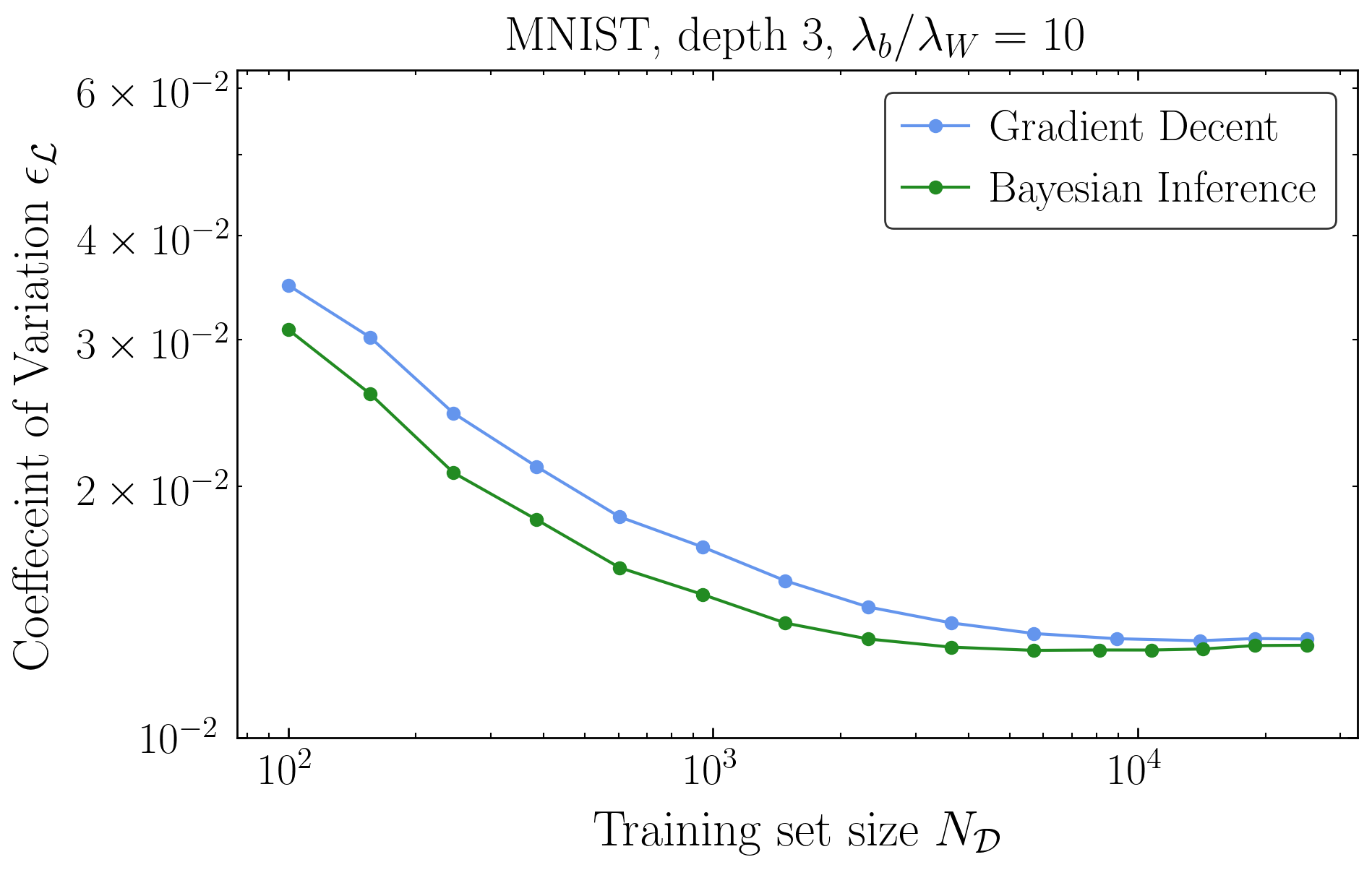}   
    \end{subfigure}
    \caption{Comparison of Bayesian inference and gradient descent scaling laws in infinite-width networks trained on MNIST. Top: mean loss $\muL$. Bottom: coefficient of variation $\relvar$.}
    \label{fig:BayesianMNIST}
\end{figure}
In physics, model uncertainty is typically expressed as the width of a Bayesian posterior, which quantifies the degree of confidence in the prediction given the training data. While neural networks trained under gradient descent do not perform exact Bayesian model fitting at finite width~\cite{Roberts:2021fes}, infinite-width MLPs \emph{can} perform Bayesian inference with a modified gradient descent algorithm where only the last layer parameters are updated~\cite{williams1996computing,neal2012bayesian,matthews2017sample,lee2018deep,de2018gaussian,lee2019wide}. The Bayesian posterior is still Gaussian, with mean and covariance given by Eqs.~(\ref{eq:infwidthpredmu})--(\ref{eq:infwidthpredsigma}) with all instances of $\boldsymbol{\Theta}$ replaced by the kernel $\mathbf{K}$. It is therefore interesting to investigate whether the infinite-width NTK scaling laws resemble those for Bayesian inference, which would imply that the infinite-width variance $\sigmaL$ (and possibly the finite-width variance as well) can be interpreted as a true Bayesian credible interval~\cite{adlamexploring}. 

In Fig.~\ref{fig:BayesianMNIST}, we show the results of this comparison for the MNIST data. In order to obtain the Bayesian prediction, the kernel has to be (numerically) invertible, which in our examples is only true for MNIST. We see that the Bayesian and NTK results for the mean agree extremely well. Furthermore, while the coefficient of variation differs somewhat between the two training algorithms at small $\ND$, both Bayesian inference and GD training asymptote to the same value of $\relvar$ at large $\ND$, with variations only at the percent level. The similarity of infinite-width Baysesian inference (also called neural network Gaussian processes) to infinite-width GD (or NTK Gaussian processes) was noted in the first papers to study the infinite-width NTK dynamics for deep networks~\cite{lee2019wide,adlamexploring}, but to our knowledge the agreement for $\relvar$ at large $\ND$ is a new result. In future work it would be interesting to investigate how robust this result is to both the depth of the network $L$ and the choice of learning rate hyperparameter $\lambda_b/\lambda_W$.

\bibliography{UQBib.bib}

\end{document}